\PassOptionsToPackage{unicode}{hyperref}
\PassOptionsToPackage{hyphens}{url}
\documentclass[10pt]{article}
\usepackage{xcolor}
\usepackage{amsmath,amssymb}
\usepackage[a4paper,margin=0.82in]{geometry}
\usepackage[section]{placeins}
\usepackage{flafter}
\usepackage{caption}
\usepackage{fancyhdr}
\usepackage{iftex}
\ifPDFTeX
  \usepackage[T1]{fontenc}
  \usepackage[utf8]{inputenc}
  \usepackage{textcomp} 
\else 
  \usepackage{unicode-math} 
  \defaultfontfeatures{Scale=MatchLowercase}
  \defaultfontfeatures[\rmfamily]{Ligatures=TeX,Scale=1}
\fi
\usepackage{lmodern}
\ifPDFTeX\else
\fi
\IfFileExists{upquote.sty}{\usepackage{upquote}}{}
\IfFileExists{microtype.sty}{
  \usepackage[]{microtype}
  \UseMicrotypeSet[protrusion]{basicmath} 
}{}
\makeatletter
\@ifundefined{KOMAClassName}{
  \IfFileExists{parskip.sty}{%
    \usepackage{parskip}
  }{
    \setlength{\parindent}{0pt}
    \setlength{\parskip}{6pt plus 2pt minus 1pt}}
}{
  \KOMAoptions{parskip=half}}
\makeatother
\usepackage{longtable,booktabs,array}
\usepackage{calc} 
\usepackage{etoolbox}
\makeatletter
\patchcmd\longtable{\par}{\if@noskipsec\mbox{}\fi\par}{}{}
\makeatother
\IfFileExists{footnotehyper.sty}{\usepackage{footnotehyper}}{\usepackage{footnote}}
\makesavenoteenv{longtable}
\usepackage{graphicx}
\usepackage{float}
\makeatletter
\newsavebox\pandoc@box
\newcommand*\pandocbounded[1]{
  \sbox\pandoc@box{#1}%
  \Gscale@div\@tempa{\textheight}{\dimexpr\ht\pandoc@box+\dp\pandoc@box\relax}%
  \Gscale@div\@tempb{\linewidth}{\wd\pandoc@box}%
  \ifdim\@tempb\p@<\@tempa\p@\let\@tempa\@tempb\fi
  \ifdim\@tempa\p@<\p@\scalebox{\@tempa}{\usebox\pandoc@box}%
  \else\usebox{\pandoc@box}%
  \fi%
}
\def\fps@figure{htbp}
\makeatother
\providecommand{\tightlist}{%
  \setlength{\itemsep}{0pt}\setlength{\parskip}{0pt}}
\usepackage[]{natbib}
\usepackage{bookmark}
\IfFileExists{xurl.sty}{\usepackage{xurl}}{} 
\hypersetup{
  pdftitle={Searching for Task-Specific Vision Paths: Evolutionary Block Pruning Across Vision-Language Models},
  pdfauthor={Tarun Tomar},
  colorlinks=true,
  linkcolor=black,
  citecolor=blue!45!black,
  urlcolor=blue!45!black,
  pdfcreator={LaTeX}}

\title{Searching for Task-Specific Vision Paths}
\usepackage{etoolbox}
\makeatletter
\providecommand{\subtitle}[1]{
  \apptocmd{\@title}{\par {\large #1 \par}}{}{}
}
\makeatother
\subtitle{Evolutionary Block Pruning Across Vision-Language Models}
\author{Tarun Tomar\\\small\url{https://github.com/TarunTomar122/vision-pathways}}
\date{}

\begin{document}
\maketitle

\begin{abstract}

Vision-language models normally execute the same complete vision encoder
for every question, even when OCR, counting, object, attribute, and
spatial queries may not require identical computation. We study whether
fixed-budget combinations of vision blocks can be skipped without
fine-tuning. A shared K-block route skips one searched set of exactly K
blocks for every question, while a capability-specific K-block policy
selects one same-size route using a known capability label. We introduce
a source-balanced evolutionary search and compare it with independent
ranking, contiguous removal, and random routes at matched budgets.
Experiments use Qwen2.5-VL-3B-Instruct, SmolVLM2-2.2B-Instruct, and an
876-example image-disjoint selection split. Search transfers across
architectures: on SmolVLM2, the searched shared four-block route beats
independent construction by 4.91 percentage points. Capability
specialization is less stable. On Qwen, the six-block capability policy
beats the shared route by 2.17 points, driven by a 7.10-point OCR gain.
On sealed IIIT5K, however, the SmolVLM2 OCR-specific route trails its
shared route by 13.6 points. Combinatorial search reliably improves
route construction, but capability labels do not define universally
transferable vision pathways.
\end{abstract}

\section{Introduction}\label{introduction}

Vision-language models (VLMs) have become a dominant paradigm for
multimodal AI. Guided by scaling laws and advances in multimodal
pretraining, VLMs have become increasingly powerful, but they have also
become expensive to deploy
\citep{kaplan2020scaling, hoffmann2022training, bai2025qwen25vl}. Unlike
language-only models, a VLM must first convert an image into visual
tokens and process them through a vision encoder before generating an
answer. Every question normally executes the same complete stack of
vision blocks, even when the visual information requested by the
question is very different.

Capabilities like OCR, counting, and spatial reasoning may not require
identical computation. This raises a simple question: does every
capability need every vision block? If some vision blocks are redundant
for a given input, replacing them with identity operations could reduce
executed vision depth without introducing sparse kernels or changing the
language decoder. However, a block that looks safe to skip by itself may
not remain safe when several blocks are skipped together because
residual blocks interact.

Existing efficiency work has explored quantization, structured layer
pruning, and visual-token reduction. Layer-pruning methods such as
ShortGPT and SliceGPT reduce depth or width in language models
\citep{men2024shortgpt, ashkboos2024slicegpt}, while DynamicViT, Token
Merging, SparseVLM, and VScan reduce the number of visual tokens
processed by transformers
\citep{rao2021dynamicvit, bolya2023tome, zhang2025sparsevlm, zhang2025vscan}.
Short-LVLM directly studies training-free layer pruning in large
vision-language models, but searches for one generally compressed model
\citep{ma2025shortlvlm}. It therefore remains unclear whether the best
removable \emph{vision-encoder} blocks depend on the requested visual
capability, and whether independently measured block importance composes
when several blocks are skipped together.

We study two questions:

\begin{enumerate}
\def\labelenumi{\arabic{enumi}.}
\tightlist
\item
  At the same number of skipped blocks, can combinatorial search find
  stronger routes than independent ranking, contiguous removal, or
  random selection?
\item
  Can capability-specific routes outperform one shared route across OCR,
  counting, object, attribute, and spatial questions?
\end{enumerate}

A route is a set of exactly \texttt{K} vision-transformer blocks
replaced by identity operations. A shared K-block route skips the same
blocks for every question, while a capability-specific K-block policy
uses one same-size route for each known capability label. We search for
the best routes found under a frozen evolutionary procedure rather than
claiming a global optimum. All comparisons use matched pruning budgets,
meaning every compared route skips exactly the same number of blocks.

\begin{figure}[t]
\centering
\includegraphics[width=0.98\linewidth]{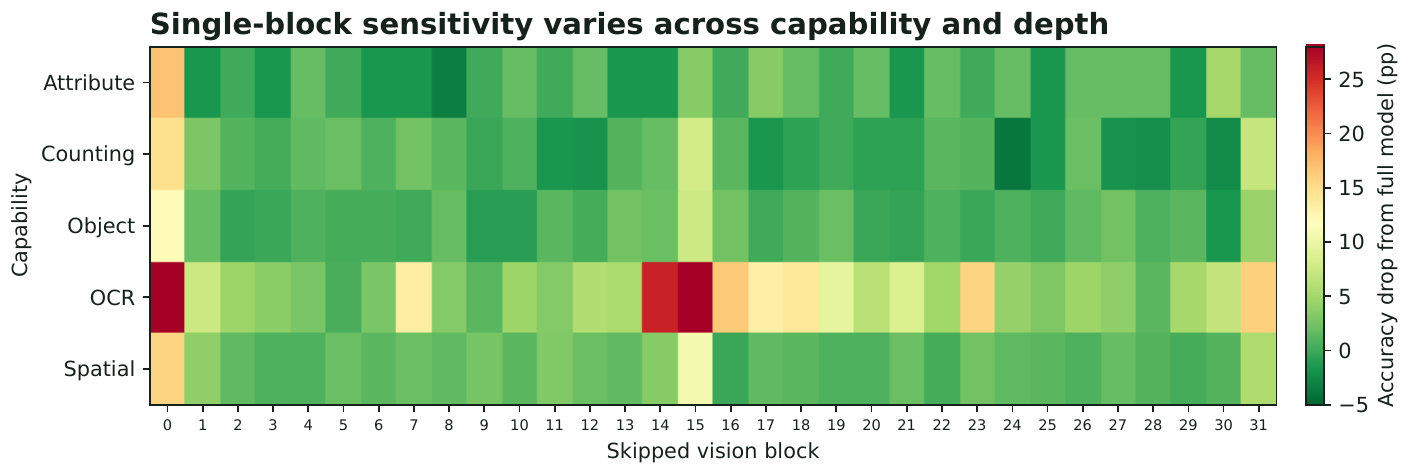}
\caption{Accuracy drop after replacing one Qwen2.5-VL-3B vision block at a time with identity. Most blocks cause limited average damage when skipped alone, but sensitivity varies by capability and depth. This is screening evidence: a low single-block drop does not imply that several such blocks can be safely combined or that a capability is stored in a specific block.}
\label{fig:single-block-sensitivity}
\end{figure}

We use Qwen2.5-VL-3B-Instruct as the primary model and
SmolVLM2-2.2B-Instruct as a second architecture. Our dataset covers five
visual capabilities: OCR, counting, object existence, attributes, and
spatial relationships. We first measure the sensitivity of every vision
block independently, then use source-balanced evolutionary search to
construct shared and capability-specific routes at matched skip budgets.
The evaluation separates route search, image-disjoint method selection,
matched controls, and a post-freeze IIIT5K source-transfer test.

We find that evolutionary search consistently constructs stronger routes
than naive block selection, but capability-specific routing is not
universally better. On Qwen, the capability-specific six-block policy is
2.17 percentage points above the shared six-block route, driven by a
7.10-point OCR gain. On SmolVLM2, evolutionary search still helps: the
searched shared four-block route beats independent construction by 4.91
points. However, its capability-specific policy is only 0.80 points
better overall because counting and spatial gains are offset by a
13.55-point OCR loss. On sealed IIIT5K, the Smol OCR-specific route
falls another 13.6 points below the shared route. These results suggest
that route construction is genuinely combinatorial, but capability
labels alone do not define stable or transferable vision pathways.

Our contributions are:

\begin{enumerate}
\def\labelenumi{\arabic{enumi}.}
\tightlist
\item
  A source-balanced evolutionary procedure for searching fixed-budget
  vision-block routes.
\item
  A matched-budget study across two VLM architectures and five visual
  capabilities.
\item
  A capability-level, cross-model, and source-transfer analysis showing
  both the potential and limitations of capability-specific routing.
\end{enumerate}

\section{Motivation}\label{motivation}

\subsection{Are individual vision blocks redundant?}\label{are-individual-vision-blocks-redundant}

Our first question was whether every vision block contributes equally to
the model's predictions. We tested this on Qwen2.5-VL-3B-Instruct, whose
vision encoder contains 32 transformer blocks \citep{bai2025qwen25vl}.
Starting from an unpruned baseline accuracy of 81.62\% on 1,480
examples, we replaced each vision block with an identity operation, one
block at a time, and measured the resulting accuracy drop. The benchmark
contains questions from five capability buckets: OCR, counting, object
existence, attributes, and spatial relationships. This produces a
block-by-capability sensitivity map rather than a pruning policy: each
intervention measures what happens when one block is removed while the
other 31 remain intact.

The result is highly uneven across depth (Figure 1). Skipping block 28
reduces overall accuracy by only 0.14 percentage points, with no
capability losing more than 1.67 points. In contrast, skipping block 15
reduces overall accuracy by 21.08 points. Blocks 0 and 31 are also
sensitive, causing drops of 17.36 and 7.77 points respectively. The
capability-level view is even less uniform. OCR accuracy falls by 25.59
points when block 14 is skipped and by 63.82 points when block 15 is
skipped, while many other block-capability pairs show little measured
damage.

These observations suggest that the vision encoder contains some
removable computation, but they do not show that a capability is stored
in a particular block. A block can look unimportant because nearby
blocks compensate for its absence, because the benchmark does not expose
all of its functions, or because its contribution only becomes necessary
after another block is removed. We therefore use the single-block sweep
only as evidence of heterogeneous sensitivity and as a source of
candidate blocks. It motivates pruning, but it cannot determine a safe
multi-block route by itself.

\subsection{Why independent block rankings do not compose}\label{why-independent-block-rankings-do-not-compose}

The most direct way to construct a deeper-pruned model is to rank blocks
by their individual accuracy drops and combine the apparently safest
ones. This follows the general intuition behind independent
layer-importance methods such as ShortGPT \citep{men2024shortgpt}. If
block effects were additive, selecting the \texttt{K} least damaging
blocks would be a reasonable approximation to the best \texttt{K}-block
route.

Our initial progressive ablation shows why this assumption is
unreliable. Skipping block 28 alone loses 0.14 points. The two-block
route \texttt{\{5,\ 28\}} loses 0.95 points, the three-block route
\texttt{\{5,\ 9,\ 28\}} loses 2.97 points, and the four-block route
\texttt{\{3,\ 5,\ 9,\ 28\}} loses 3.72 points. The final route removes
12.5\% of the vision blocks and improves measured vision-encoder latency
by 4.89\%, but it loses 7.94 points on OCR. Blocks that appear safe
under separate interventions can therefore become damaging when
composed.

This failure is expected in a sequential residual network. Each block
receives the representation produced by all preceding blocks, so
removing one block changes the input distribution seen by every block
after it. Let \(\Delta(S)\) denote the accuracy loss after skipping a
set of blocks \(S\). For two blocks \(i\) and \(j\), we can describe
their pairwise interaction as

\[
I(i,j) = \Delta(i,j) - \Delta(i) - \Delta(j).
\]

A positive value means that the pair is more harmful than their
independent effects predict, while a negative value means that the
blocks are more compatible to skip together. Higher-order routes can
contain interactions that no collection of one-block scores can recover.
Block importance is therefore conditional on which other blocks are
already skipped. The object we need to select is not a list of blocks,
but a complete route.

\subsection{Why route selection is a combinatorial problem}\label{why-route-selection-is-a-combinatorial-problem}

Once route quality depends on block combinations, evaluating blocks
independently is insufficient. For a vision encoder with \(L\) blocks
and a fixed skip budget \(K\), the number of possible routes is

\[
|\Omega_K| = \binom{L}{K}.
\]

For Qwen's 32-block vision encoder, this gives 35,960 possible
four-block routes, 906,192 six-block routes, and 10,518,300 eight-block
routes. Exhaustively running every route through a VLM benchmark would
require millions of full inference evaluations at the larger budgets.
The search space becomes even more expensive if a separate route is
considered for each capability.

This motivates a fixed-budget combinatorial search. The method must
compare routes with the same value of \(K\), evaluate interactions among
skipped blocks, and balance average performance against severe failures
on individual data sources. It must also support two distinct
hypotheses: one shared route may work well for all questions, or
different capability labels may favor different routes of the same size.
Finally, searched routes must be compared against independent rankings,
contiguous removals, and random routes under exactly matched skip
budgets. These requirements lead us to the source-balanced evolutionary
search described next.

\section{Methodology}\label{methodology}

\subsection{Identity block skipping}\label{identity-block-skipping}

Let a frozen vision encoder contain \(L\) transformer blocks
\(F_0,\ldots,F_{L-1}\). We represent a route by a subset of block
indices

\[
S \subseteq \{0,\ldots,L-1\}, \qquad |S|=K,
\]

where \(S\) contains the blocks to skip and \(K\) is the skip budget.
For hidden state \(h_l\), executing route \(S\) changes the vision
encoder to

\[
h_{l+1} =
\begin{cases}
h_l, & l \in S,\\
F_l(h_l), & l \notin S.
\end{cases}
\]

Selected blocks are therefore replaced by identity operations. We do not
remove the patch embedding, token merger, language decoder, or any block
parameters from the checkpoint, and we do not fine-tune the model. This
intervention measures the effect of reducing executed vision depth while
leaving the rest of the inference pipeline unchanged. Within each
comparison, every route contains exactly \(K\) unique block indices.

\subsection{Shared and capability-specific routes}\label{shared-and-capability-specific-routes}

We study two route families. A \textbf{shared route} uses one set
\(S_{\mathrm{shared}}\) for every question, regardless of its
capability. This is the standard generic pruning setting: one fixed
compressed pathway must preserve average behavior across the full
benchmark.

A \textbf{capability-specific policy} contains one route \(S_t\) for
each capability in the set
\[
\mathcal{T} = \{\text{attribute},\ \text{counting},\ \text{object},\
\ \text{OCR},\ \text{spatial}\}.
\]
At
inference, the route associated with the known capability label is
selected before the image passes through the vision encoder. Every
\(S_t\) has the same skip budget \(K\), so a policy cannot gain accuracy
by executing more blocks for a difficult capability. This setup tests
whether different capabilities prefer different fixed block subsets.

The policy is an oracle-label evaluation, not a learned router. Our
experiments assume the capability label is already available and do not
measure classification errors or routing overhead. The search also
returns one frozen route per label rather than choosing a new route for
every image or question.

\subsection{Source-balanced objectives}\label{source-balanced-objectives}

Raw benchmark accuracy can hide source imbalance because some datasets
contribute many more questions than others. We therefore evaluate each
route against the full model on exactly paired examples and aggregate at
the capability-source level. Let \(I_{c,d}\) be the examples belonging
to capability \(c\) and dataset source \(d\). If \(y_i(S)\) is one when
route \(S\) answers example \(i\) correctly and zero otherwise, its cell
accuracy is

\[
A^S_{c,d}=\frac{1}{|I_{c,d}|}\sum_{i\in I_{c,d}} y_i(S).
\]

The paired damage relative to the full model is measured in percentage
points:

\[
\Delta_{c,d}(S)=100\left(A^0_{c,d}-A^S_{c,d}\right).
\]

Positive values indicate lost accuracy, while negative values mean that
the skipped model happened to outperform the full model in that cell.
Every capability-source cell receives equal weight regardless of its
number of examples. Failed or malformed model outputs count as
incorrect.

For a shared route, let \(G\) be all capability-source cells and define
its mean damage, worst-source damage, and source variability as

\[
\mu(S)=\frac{1}{|G|}\sum_{g\in G}\Delta_g(S), \qquad
w(S)=\max_{g\in G}\Delta_g(S),
\]

\[
\sigma(S)=\sqrt{\frac{1}{|G|}\sum_{g\in G}\left(\Delta_g(S)-\mu(S)\right)^2}.
\]

Evolutionary survival minimizes the vector
\(q_{\mathrm{shared}}(S)=(\mu,w,\sigma)\). When finalists must be
ordered, we use the frozen scalar loss

\[
L_{\mathrm{shared}}(S)=0.50\mu(S)+0.30w(S)+0.20\sigma(S).
\]

For a capability-specific route targeting capability \(t\), let \(T_t\)
contain its source cells and let \(C_t=G\setminus T_t\) contain all
non-target cells. We define

\[
\mu_t(S)=\frac{1}{|T_t|}\sum_{g\in T_t}\Delta_g(S), \qquad
w_t(S)=\max_{g\in T_t}\Delta_g(S),
\]

\[
\sigma_t(S)=\sqrt{\frac{1}{|T_t|}\sum_{g\in T_t}\left(\Delta_g(S)-\mu_t(S)\right)^2},
\]

and measure mean collateral damage as

\[
\kappa_t(S)=\frac{1}{|C_t|}\sum_{g\in C_t}\Delta_g(S).
\]

Evolutionary survival minimizes
\(q_t(S)=(\mu_t,w_t,\kappa_t,\sigma_t)\), and finalist ranking uses

\[
L_t(S)=0.45\mu_t(S)+0.30w_t(S)+0.15\kappa_t(S)+0.10\sigma_t(S).
\]

The collateral term discourages a route from preserving its target by
destroying every other capability. It remains a soft penalty rather than
a hard constraint. The worst-source and variability terms similarly
prevent a good mean from concealing severe damage on one source.

\subsection{Evolutionary route search}\label{evolutionary-route-search}

The search space contains fixed-size block subsets, so we use a seeded
genetic algorithm with NSGA-II-style multi-objective survival
\citep{deb2002nsga2}. Each route family and skip budget is searched
independently. The procedure is summarized in Figure 2.

\begin{figure}[t]
\centering
\includegraphics[width=0.98\linewidth]{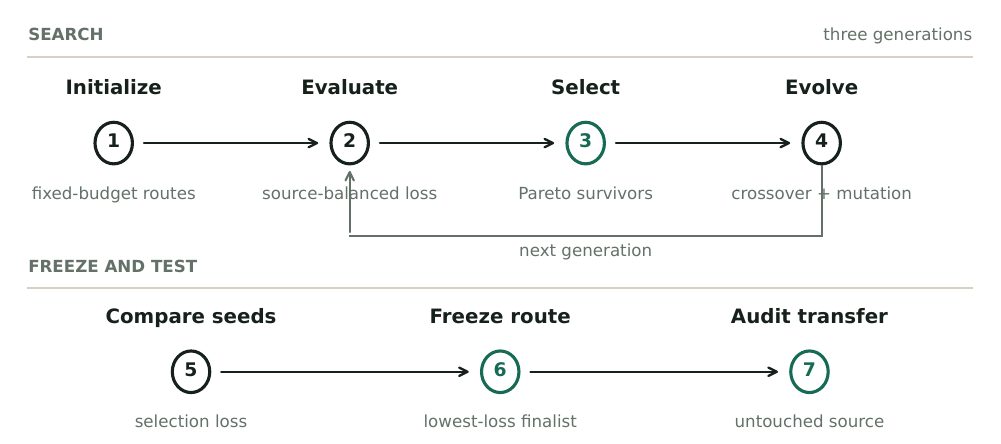}
\caption{Fixed-budget evolutionary route search. Candidate routes are evaluated with paired source-aware objectives, filtered by Pareto survival, evolved through crossover and mutation, and frozen only after three-seed finalist selection. Held-out transfer is an audit after the route is frozen.}
\label{fig:method-overview}
\end{figure}

\textbf{Initialization.} At most half of each population is filled with
prior routes constructed from the least-damaging single-block rankings
and earlier combined-route searches, resized to the current value of
\(K\). The remaining positions are unique, seeded random \(K\)-subsets
of the allowed vision blocks. Priors improve search efficiency but do
not restrict which blocks later mutation can introduce.

\textbf{Pareto survival.} A candidate \(S_a\) dominates \(S_b\) when it
is no worse on every component of the relevant objective vector and
strictly better on at least one:

\[
S_a \prec S_b \iff
\left[\forall j,\ q_j(S_a)\le q_j(S_b)\right]
\land
\left[\exists j,\ q_j(S_a)<q_j(S_b)\right].
\]

Candidates are divided into nondominated fronts, which are accepted in
order until half the population survives. If only part of a front fits,
we prefer higher NSGA-II crowding distance, then lower scalar loss, then
lexicographically smaller routes. Survivors are copied unchanged into
the next generation. Before reproduction they are ordered by scalar loss
and route index. For ordered survivors \(R_0,\ldots,R_{m-1}\), offspring
attempt \(i\) pairs

\[
A_i=R_{i\bmod m}, \qquad B_i=R_{(5i+1)\bmod m}.
\]

This deterministic schedule replaces stochastic parent selection.

\textbf{Fixed-\(K\) crossover.} Given two same-size parent routes \(A\)
and \(B\), the child retains every block in their intersection and fills
its remaining positions from their non-shared union:

\[
C=(A\cap B)\cup U, \qquad
U\subseteq(A\cup B)\setminus(A\cap B), \qquad
|U|=K-|A\cap B|.
\]

The subset \(U\) is chosen by seeded SHA-256 ordering. This preserves
shared parental choices while guaranteeing that the child still skips
exactly \(K\) blocks.

\textbf{One-swap mutation.} Every crossed child is mutated exactly once
by removing one included block \(r\) and adding one allowed block \(a\)
outside the child:

\[
M(C)=(C\setminus\{r\})\cup\{a\}.
\]

The removal and addition are again selected by seeded SHA-256 ordering.
Duplicate children are rejected, and a deterministic random route is
inserted only if repeated crossover-mutation attempts cannot refill the
population. There is no crossover or mutation probability: every
offspring receives both operations once, while elite survivors receive
neither. Fitness acts through Pareto survival and deterministic survivor
ordering rather than roulette-wheel or tournament selection.

\subsection{Finalist selection and matched controls}\label{finalist-selection-and-matched-controls}

We run each route-family and budget search with three fixed seeds:
20260715, 20260716, and 20260717. Qwen uses a population of 16 routes
and three evaluated generations, including the initial population, for
\(K\in\{4,6,8\}\). Each seed contributes at most two Pareto finalists.
After deduplication across seeds, all finalists are evaluated on a
300-example development set. At most three routes with the lowest
matching scalar loss advance to the 876-example image-disjoint selection
set, and the route with the lowest selection loss is frozen. Exact ties
are broken by lexicographic block order.

The SmolVLM2 replication follows the same objectives, operators, three
seeds, and finalist procedure, but uses a population of 12, two
evaluated generations, and only \(K=4\). These leaner settings were
frozen before the replication run. Routes are never automatically reused
across budgets or architectures.

Every final comparison is made at matched \(K\). The controls are: the
\(K\) blocks with the lowest independent average sensitivity, the \(K\)
blocks with the lowest capability-specific sensitivity, the
least-damaging contiguous \(K\)-block window, and three deterministic
random \(K\)-block routes. The unpruned model is reported as an accuracy
reference, not as a matched-compute control. The 876-example set is used
for method selection and is not described as sealed evidence because
earlier discovery work touched its underlying benchmark sources.
External transfer data is accessed only after routes, objectives,
prompts, decoding, and scoring rules are frozen.

We quantify uncertainty with 10,000 fixed-seed paired bootstrap
resamples over the same example IDs. Reported intervals are percentile
95\% intervals for the accuracy difference between two routes. We treat
an interval that crosses zero as inconclusive; an interval that only
touches zero is reported as borderline rather than as a clear positive
result.

\section{Experiments}\label{experiments}

\subsection{Models, datasets, and metrics}\label{models-datasets-and-metrics}

\textbf{Models.} The primary model is Qwen2.5-VL-3B-Instruct, which
contains 32 vision blocks \citep{bai2025qwen25vl}. The replication model
is SmolVLM2-2.2B-Instruct, which contains 27 vision blocks
\citep{huggingface2025smolvlm2}. We pin Qwen revision
\nolinkurl{66285546d2b821cf421d4f5eb2576359d3770cd3} and SmolVLM2 revision
\nolinkurl{482adb537c021c86670beed01cd58990d01e72e4}. Both models use
deterministic greedy decoding and batch-size-one inference. The
full-model accuracies on the 876-example selection partition are 83.68\%
for Qwen and 82.65\% for SmolVLM2.

\textbf{Data.} The discovery corpus contains 1,780 image-question pairs
over 1,431 unique images. It combines controlled MME perception
questions \citep{fu2023mme} with one larger source for each capability:
OCRBench for direct text recognition \citep{liu2023ocrbench}, TallyQA
for counting \citep{acharya2019tallyqa}, POPE for object existence
\citep{li2023pope}, VQAv2 color questions for attributes
\citep{goyal2017vqav2}, and VSR for spatial relationships
\citep{liu2023vsr}. Images are assigned as complete hash groups,
producing 904 development examples and an image-disjoint 876-example
selection partition. The selection partition contains 166 attribute, 181
counting, 193 object, 155 OCR, and 181 spatial questions.

The development and selection partitions are separated by image, but
both source families participated in earlier single-block discovery. We
therefore use the 876 examples as method-selection evidence rather than
calling them a sealed test set. A separate 250-example IIIT5K scene-text
subset \citep{mishra2012iiit5k} is kept sealed until the SmolVLM2 shared
and OCR-specific four-block routes are frozen. This final set tests
source transfer only; it does not influence route selection, objectives,
prompts, decoding, or scoring.

\textbf{Metrics.} The primary metric is short-answer accuracy, reported
overall and within each capability. Route comparisons use paired
accuracy differences in percentage points with the bootstrap intervals
defined in Section 3.5. We also report mean pairwise Jaccard overlap
among the three seed-level route winners, the fraction of vision and
total-model parameters bypassed by identity skipping, and batch-size-one
vision-encoder and end-to-end latency. Parameter counts describe
inactive execution, not the size of the saved checkpoint.

\subsection{Does evolutionary search beat naive pruning?}\label{does-evolutionary-search-beat-naive-pruning}

Figure 3 compares route construction methods across Qwen skip budgets.
The searched shared route achieves 81.28\%, 79.11\%, and 75.91\%
accuracy when four, six, and eight blocks are skipped. Independent
ranking reaches 80.25\%, 78.08\%, and 72.37\%, so evolutionary search
improves accuracy by 1.03, 1.03, and 3.54 points respectively. The
advantage becomes much larger against weaker controls at aggressive
budgets. With eight blocks skipped, contiguous removal reaches 40.98\%
and the mean of three random routes reaches 46.96\%, compared with
75.91\% for the searched shared route.

\begin{figure}[t]
\centering
\includegraphics[width=0.98\linewidth]{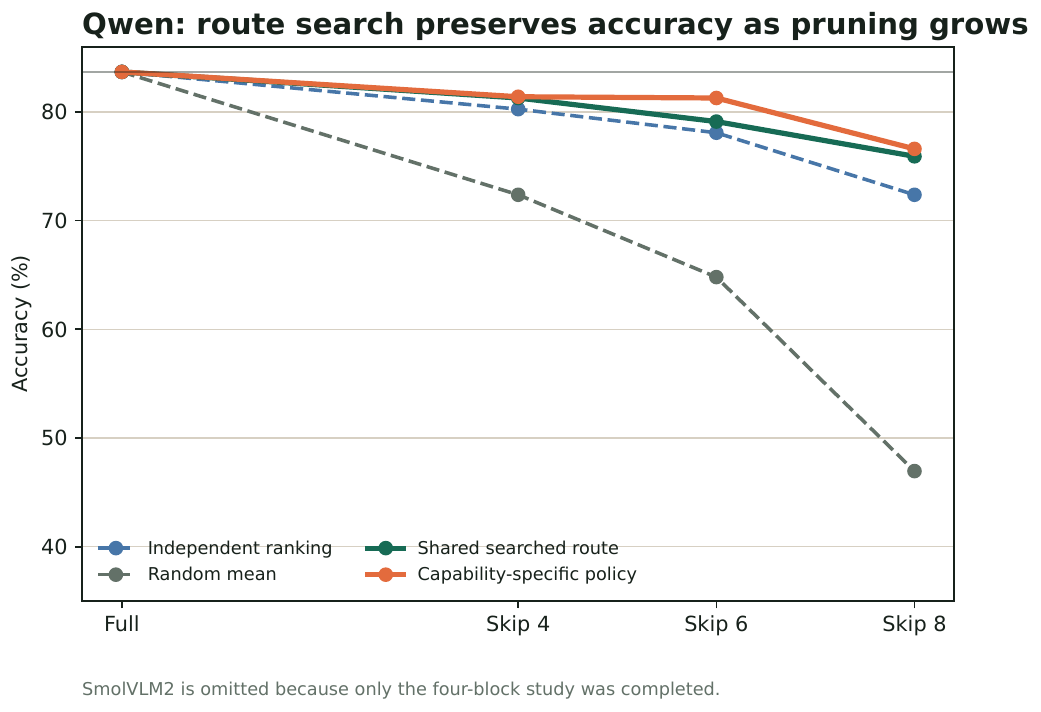}
\caption{Qwen selection accuracy as the number of skipped blocks increases. Every condition at a given x-axis position skips exactly the same number of vision blocks. SmolVLM2 is omitted because only its four-block study was completed.}
\label{fig:qwen-budget}
\end{figure}

The same pattern appears for capability-specific construction. On Qwen,
the searched capability policy exceeds independently constructed
capability routes by 0.11 points with four blocks skipped, 3.77 points
with six skipped, and 8.45 points with eight skipped. This increasing
gap supports the central motivation: one-block scores become less
reliable as more interventions must compose.

The four-block comparison replicates on SmolVLM2 (Figure 4). Its
searched shared route reaches 72.49\%, compared with 67.58\% for
independent ranking, a gain of 4.91 points with paired 95\% interval
{[}1.83, 7.99{]}. Its searched capability policy reaches 73.29\%,
compared with 64.38\% for independently constructed capability routes, a
gain of 8.90 points {[}5.82, 11.99{]}. The contiguous route reaches
71.58\% and the random mean reaches 54.53\%. Evolutionary route
construction therefore improves same-budget pruning on both
architectures, even though their absolute robustness differs.

\begin{figure}[t]
\centering
\includegraphics[width=0.98\linewidth]{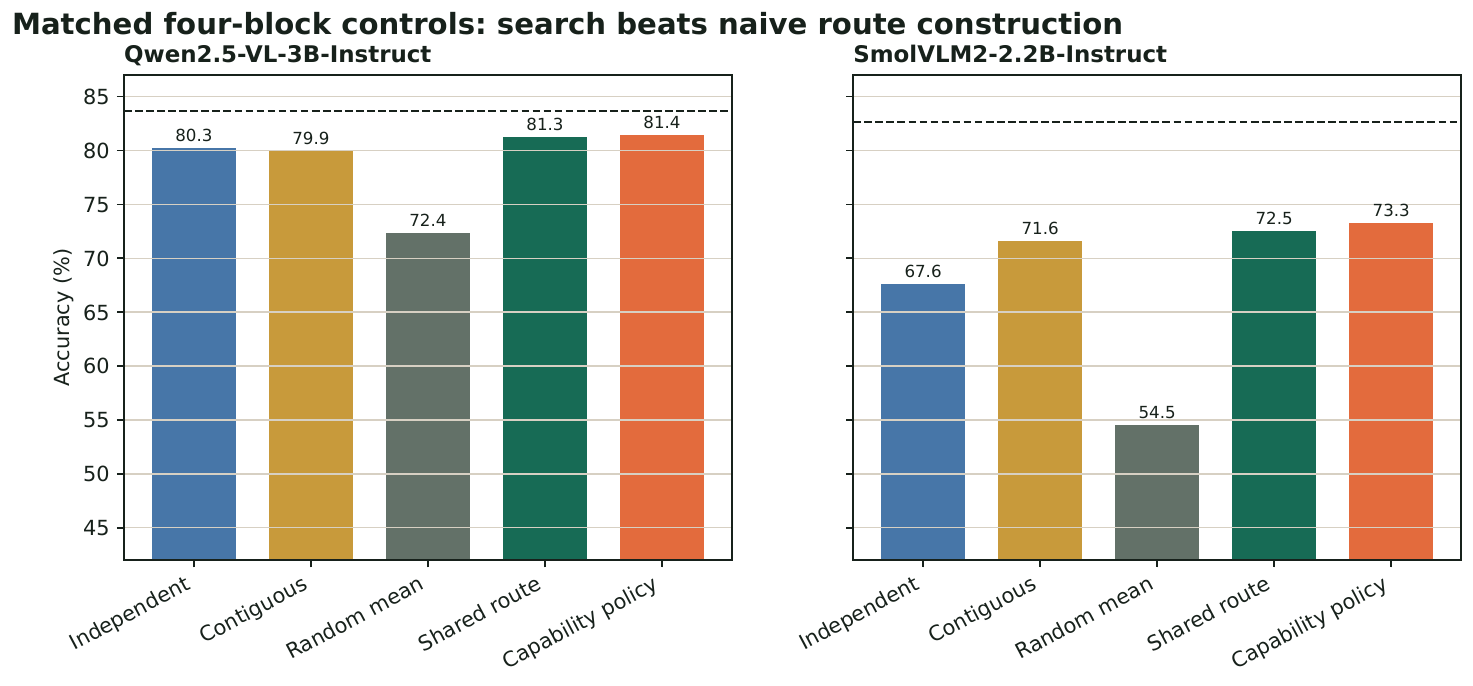}
\caption{Accuracy for matched four-block controls. Dashed lines indicate the corresponding full-model accuracy. Search improves route construction on both architectures, while SmolVLM2 remains more damaged by four skipped blocks overall.}
\label{fig:matched-k4}
\end{figure}

\subsection{Does capability-specific routing help?}\label{does-capability-specific-routing-help}

Table 1 separates the search benefit from the stronger claim that
capability-specific policies outperform one shared route. With four Qwen
blocks skipped, the shared route and capability policy score 81.28\% and
81.39\%. The difference is only +0.11 points, and its interval {[}-1.83,
2.05{]} crosses zero. With eight blocks skipped, the policy gains +0.68
points {[}-1.71, 3.08{]}, which is also inconclusive.

\begin{table}[t]
\centering
\caption{Matched-budget selection accuracy. A shared route uses one block set for every question; a capability-specific policy uses one same-size route per known capability label.}
\label{tab:main-results}
\resizebox{\linewidth}{!}{\begin{tabular}{llrrrrl}
\toprule
Model & Skipped & Full & Shared & Capability & $\Delta$ (pp) & 95\% CI \\
\midrule
Qwen2.5-VL-3B & 4 blocks & 83.68 & 81.28 & 81.39 & +0.11 & [-1.83, 2.05] \\
Qwen2.5-VL-3B & 6 blocks & 83.68 & 79.11 & 81.28 & +2.17 & [0.00, 4.34] \\
Qwen2.5-VL-3B & 8 blocks & 83.68 & 75.91 & 76.60 & +0.68 & [-1.71, 3.08] \\
SmolVLM2-2.2B & 4 blocks & 82.65 & 72.49 & 73.29 & +0.80 & [-1.94, 3.54] \\
\bottomrule
\end{tabular}
}
\end{table}

The strongest Qwen specialization result occurs with six blocks skipped.
The capability policy reaches 81.28\%, compared with 79.11\% for the
shared route, for a +2.17-point difference {[}0.00, 4.34{]}. The
interval touches zero, so the overall result is borderline rather than
unambiguously positive. At the capability level, OCR provides clearer
evidence: its selected route gains 7.10 points {[}0.65, 14.19{]} over
the shared route. Attribute gains 1.20 points, counting 1.10, spatial
2.21, and object is unchanged, but all four intervals include zero. Qwen
therefore supplies evidence for useful specialization at one budget and
capability, not a general rule that every capability needs a different
route.

\subsection{Does the result replicate across models?}\label{does-the-result-replicate-across-models}

SmolVLM2 replicates the value of evolutionary search but not the Qwen
capability profile. Its four-block capability policy is only 0.80 points
above its shared route overall, with interval {[}-1.94, 3.54{]}. That
small aggregate difference hides large effects in opposite directions
(Figure 5). Counting improves by 7.18 points {[}1.10, 13.26{]} and
spatial reasoning improves by 9.39 points {[}2.76, 16.02{]}. OCR moves
in the opposite direction, falling by 13.55 points {[}-21.94, -5.16{]}.
Attribute changes by -0.60 points and object by -0.52, with both
intervals crossing zero.

\begin{figure}[t]
\centering
\includegraphics[width=0.98\linewidth]{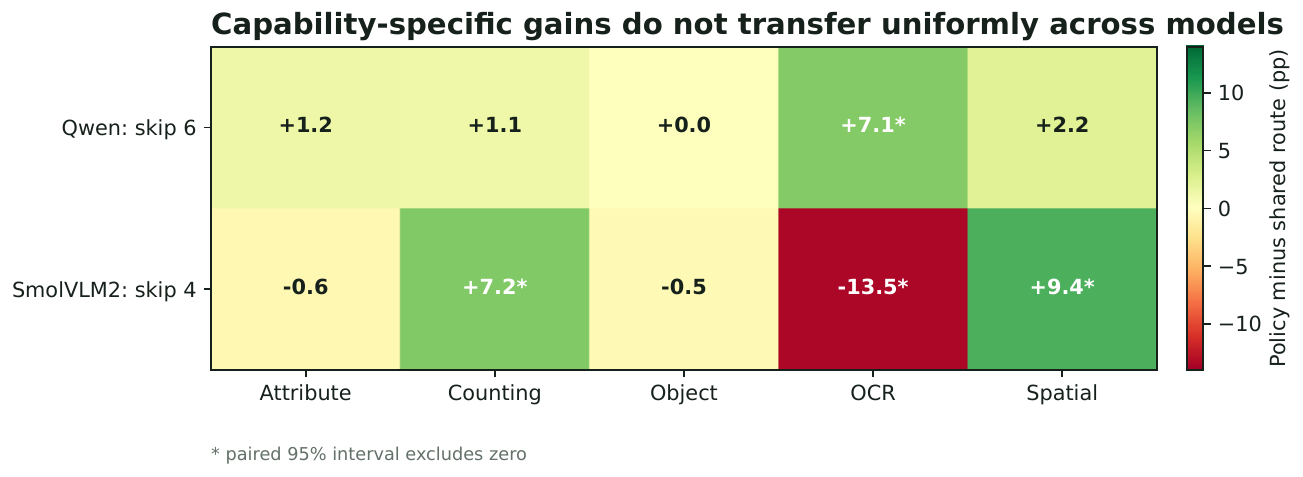}
\caption{Capability-policy accuracy minus shared-route accuracy. Stars mark paired 95\% intervals that exclude zero. Qwen is shown at its strongest specialization budget, while SmolVLM2 is shown at its completed four-block budget.}
\label{fig:cross-model-capability}
\end{figure}

The sign reversal for OCR is especially important. A capability-specific
OCR route helps Qwen at six skipped blocks but harms SmolVLM2 at four.
Similar aggregate policy scores can therefore conceal fundamentally
different behavior by capability. The result argues against interpreting
one successful route as a universal capability-to-depth map. Plausible
explanations include architecture-specific block interactions,
source-specific shortcuts, and noise from a finite search budget; this
experiment cannot isolate which explanation dominates.

\subsection{Does an OCR-specific route transfer to IIIT5K?}\label{does-an-ocr-specific-route-transfer-to-iiit5k}

We test the frozen SmolVLM2 routes on 250 IIIT5K word images that were
never accessed during route search. The full model reaches 94.8\%
exact-match accuracy. The searched shared four-block route reaches
86.4\%, while the OCR-specific four-block route reaches only 72.8\%
(Figure 6). The route selected specifically for OCR is therefore 13.6
points worse than the shared route, with paired 95\% interval {[}-19.2,
-8.4{]}.

\begin{figure}[t]
\centering
\includegraphics[width=0.98\linewidth]{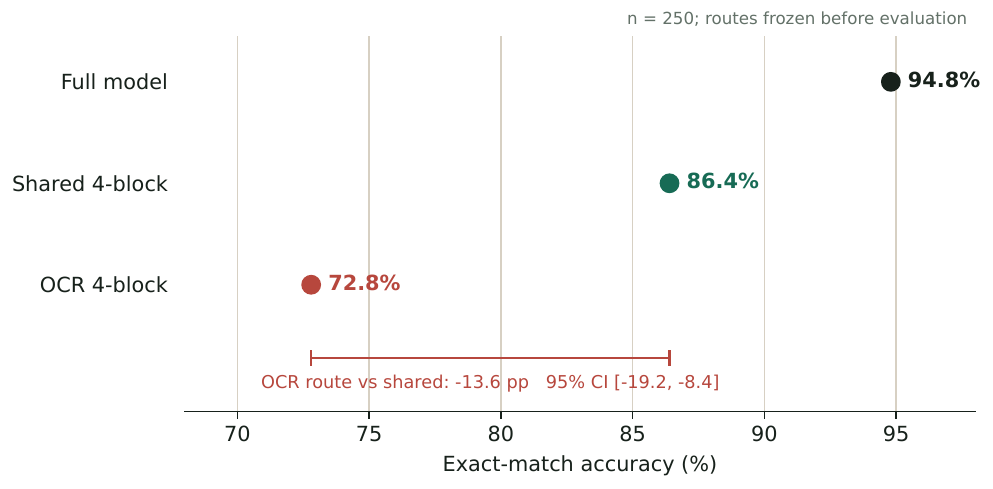}
\caption{Post-freeze source transfer on IIIT5K. Both compressed conditions skip four SmolVLM2 vision blocks. Routes and all evaluation choices were frozen before any IIIT5K prediction was produced.}
\label{fig:iiit5k-transfer}
\end{figure}

This negative transfer rules out the strongest interpretation of the
capability-specific hypothesis. The selected SmolVLM2 OCR route is not a
stable OCR pathway across sources. It may have adapted to the mixture of
MME and OCRBench examples used during selection, exploited
architecture-specific interactions that do not support IIIT5K text, or
won because several low-stability candidates had similar development
losses. The experiment demonstrates the failure but cannot distinguish
these mechanisms. It also shows why a capability name alone may be too
coarse for routing: source, text style, image complexity, and prompt
format may matter alongside the nominal task.

\subsection{Efficiency and route stability}\label{efficiency-and-route-stability}

The selected block identities are often unstable across search seeds.
Figure 7 reports the mean pairwise Jaccard overlap among three
development winners at the four-block budget. For the shared route,
overlap is 0.095 on Qwen and 0.159 on SmolVLM2. Qwen's attribute and OCR
searches are more stable at 0.733 and 0.556, but its spatial winners
have zero overlap. Every SmolVLM2 route family remains at or below
0.222. Different block sets can therefore produce similar objective
values, which further weakens any claim that the search recovers one
canonical pathway for a capability.

\begin{figure}[t]
\centering
\includegraphics[width=0.98\linewidth]{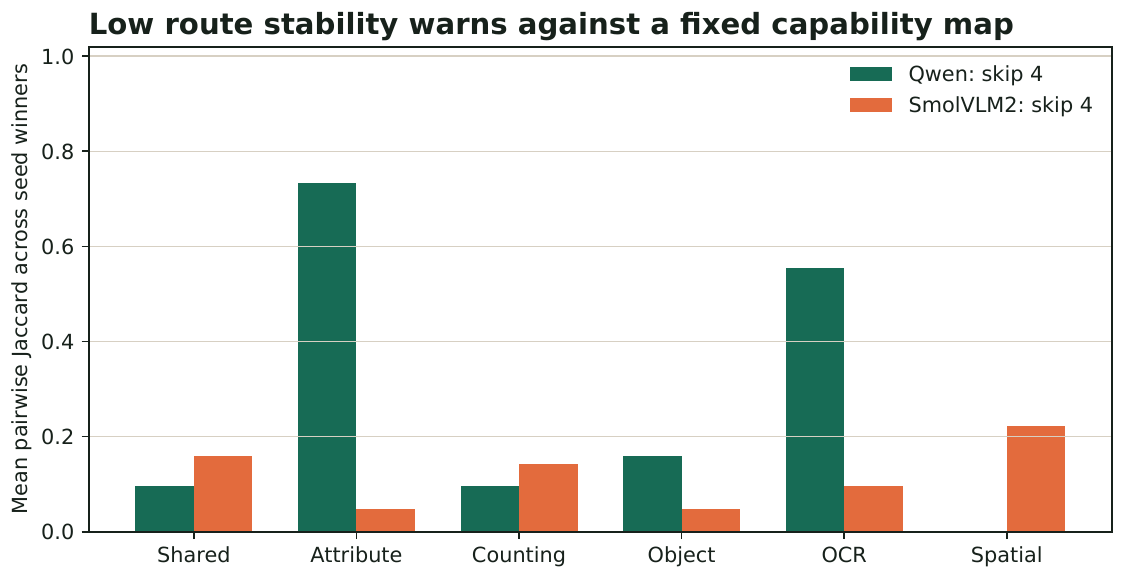}
\caption{Mean pairwise Jaccard overlap among three seed-level four-block development winners. Low overlap indicates that the frozen route is one strong solution among several distinct candidates, not a uniquely identified block map.}
\label{fig:route-stability}
\end{figure}

\FloatBarrier

Identity skipping produces a meaningful reduction in executed vision
depth but a modest reduction in whole-model parameters (Figure 8). Four
Qwen blocks account for 11.79\% of vision-tower parameters and 2.10\% of
total model parameters. Four SmolVLM2 blocks account for 14.76\% of
vision parameters and 2.71\% of total parameters. These values describe
parameters bypassed during execution. The original modules remain in the
loaded checkpoint, so our implementation does not yet produce a
physically smaller model file.

\begin{figure}[H]
\centering
\includegraphics[width=0.98\linewidth]{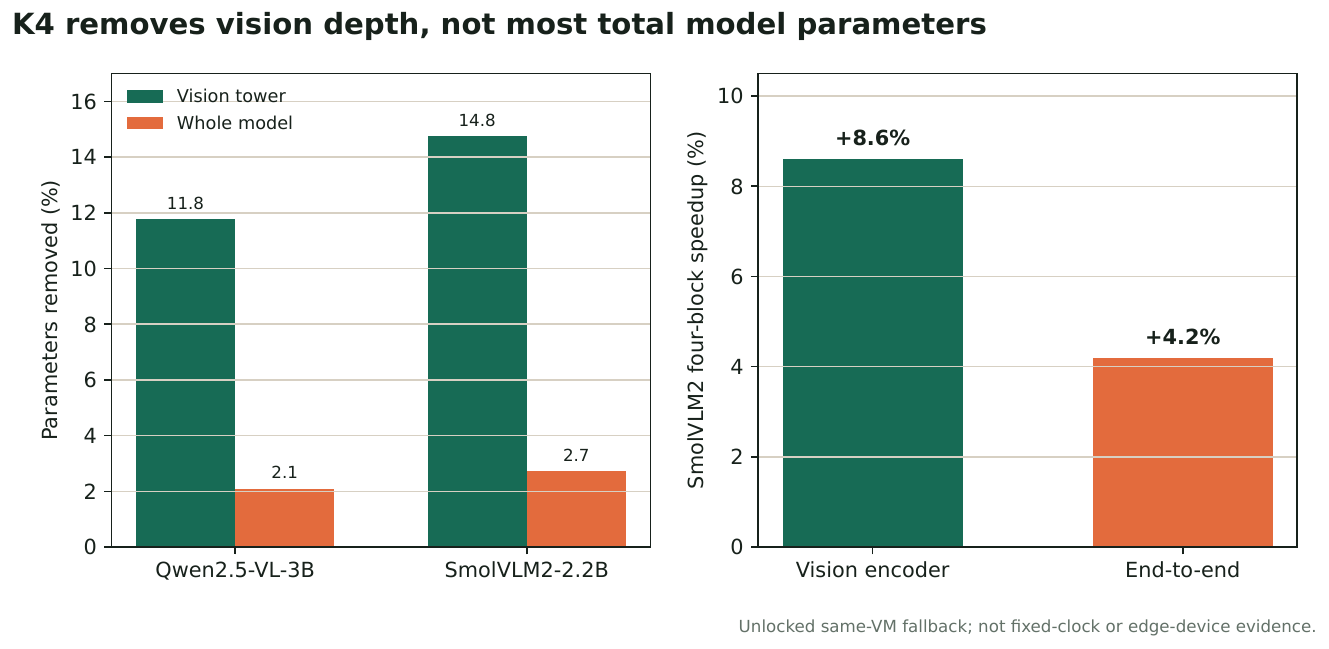}
\caption{Four-block parameter fractions for both models and measured SmolVLM2 latency. The latency measurement is an unlocked same-VM comparison on one RTX 4090, not fixed-clock or edge-device evidence.}
\label{fig:efficiency}
\end{figure}

For the searched SmolVLM2 shared four-block route, repeated
batch-size-one measurements show an 8.60\% vision-encoder speedup and a
4.19\% end-to-end speedup on the same RTX 4090 VM. The smaller
end-to-end gain is expected because image preprocessing and
autoregressive language generation are unchanged. We do not report a
final Qwen accuracy-latency curve: its committed latency audit predates
the final evolved routes and does not provide matched measurements
across the four-, six-, and eight-block winners.

Together, the results support four conclusions. First, block
interactions make route construction combinatorial, and evolutionary
search is consistently better than naive composition. Second, a safe
pruning budget is architecture-dependent: SmolVLM2 loses 10.16 points
from its full model even under its searched shared four-block route,
while Qwen loses 2.40. Third, aggregate accuracy can hide severe
capability-level gains and losses. Fourth, capability labels alone are
not reliable routing variables. A learned router conditioned on
question, source, image complexity, and hardware cost is a promising
extension, but it is not evaluated here.

\section{Limitations}\label{limitations}

This study covers two model architectures, five visual capability
labels, and one constructed mixture of short-answer benchmarks. The
evolutionary budget is finite, so every selected route is the best found
under the frozen search rather than a globally optimal subset. Dataset
identity also remains partly confounded with capability. Although MME
provides a second controlled source within each bucket and the objective
balances capability-source cells, OCRBench, TallyQA, POPE, VQAv2, and
VSR still contribute different image and answer distributions. Public
benchmark examples may also have appeared in model pretraining data.

The 876-example selection partition is image-disjoint from route
development, but it is not a pristine held-out benchmark because its
source families participated in earlier single-block screening. Only the
250-example IIIT5K audit is a genuinely fresh post-freeze
source-transfer test, and it evaluates one capability on one model. The
paired bootstrap intervals quantify uncertainty over the observed
examples. They do not capture uncertainty across model training runs,
benchmark families, search hyperparameters, or alternative capability
taxonomies.

Our capability-specific policy assumes an oracle capability label. We do
not train a per-question router, measure routing mistakes, or include
routing overhead. Identity skipping updates no weights, so the study
does not test recovery through fine-tuning, distillation, or low-rank
adaptation. It also evaluates deterministic short-answer correctness
rather than long-form generation, grounding, calibration, or safety
behavior.

Finally, our efficiency evidence is limited. The implementation bypasses
blocks at runtime but retains their parameters in the checkpoint. It
does not produce a deployable compact checkpoint, prune the language
decoder, quantize weights, or combine depth skipping with visual-token
reduction. Latency was measured at batch size one on one RTX 4090 VPS.
The SmolVLM2 result is an unlocked same-VM comparison because fixed GPU
clocks were unavailable, and no mobile or edge hardware, memory
bandwidth, energy consumption, or complete Qwen final-route latency
series was measured.

\section{Related Work}\label{related-work}

\textbf{Structured depth and width reduction.} ShortGPT measures
residual-block influence and removes low-influence language-model layers
without retraining \citep{men2024shortgpt}. SliceGPT exploits
computational invariance to delete rows and columns from transformer
weight matrices, reducing hidden width rather than depth
\citep{ashkboos2024slicegpt}. These methods establish that structured
compression can map to dense hardware operations, but neither asks
whether combinations of vision-encoder blocks should vary by visual
capability.

Short-LVLM directly studies training-free layer pruning in large
vision-language models \citep{ma2025shortlvlm}. It uses important
vision-language tokens to improve layer localization and applies a
low-rank subspace compensation to reduce the feature gap after pruning.
Its objective is a generally compressed model, whereas our primary
empirical question compares one shared route with a capability-specific
policy and tests whether independently safe vision blocks compose at
matched skip budgets. The methods are complementary: compensation could
be applied after our route search, while our results caution that
localization quality alone may not predict multi-block compatibility.

\textbf{Visual-token reduction.} DynamicViT learns input-dependent token
sparsification within vision transformers \citep{rao2021dynamicvit},
while Token Merging combines similar tokens without retraining
\citep{bolya2023tome}. SparseVLM and VScan reduce visual-token
computation in large vision-language models
\citep{zhang2025sparsevlm, zhang2025vscan}, and FlowCut analyzes
information flow to identify redundant visual computation
\citep{tong2025flowcut}. These approaches primarily reduce sequence
width or token processing. Our intervention reduces executed
vision-encoder depth, making route search potentially compatible with
token reduction rather than a replacement for it.

\textbf{Conditional VLM pruning.} FlashVLM uses visual attention to
guide inference-time layer skipping during VLM decoding
\citep{sarkar2026flashvlm}. Domain-aware pruning work studies how
decoder-layer rankings change across domains
\citep{khaki2026domainaware}. This work shares the broad idea that
redundancy can depend on the input distribution, but differs in
intervention target and evaluation. We search fixed-cardinality subsets
in the vision tower, compare shared and named-capability routes at
identical budgets, and explicitly test cross-architecture behavior,
source transfer, and seed stability. Our findings do not imply that
task-aware VLM efficiency is otherwise unexplored; they show that
capability-conditioned vision-depth routes are less stable than
aggregate method-selection accuracy suggests.

\section{Conclusion}\label{conclusion}

We asked whether combinations of vision-encoder blocks can be skipped
more intelligently than independent layer ranking, and whether different
visual capabilities benefit from different routes. Across Qwen2.5-VL-3B
and SmolVLM2-2.2B, the first answer is consistently yes: source-balanced
evolutionary search constructs stronger same-budget routes than
independent, contiguous, or random selection. The method is especially
valuable as the skip budget grows and block interactions make one-block
rankings unreliable.

The second answer is conditional. Qwen shows a useful six-block
capability policy, including a clear OCR gain, while SmolVLM2 shows
large counting and spatial gains alongside a larger OCR loss. The sealed
IIIT5K test confirms that the SmolVLM2 OCR-specific route does not
transfer even within the capability it was designed to preserve. We
therefore do not find a stable capability-to-layer map. The practical
result is narrower but reproducible: combinatorial search is a useful
tool for vision-depth pruning, while reliable dynamic routing will
require richer per-input signals, stronger multi-source validation, and
hardware-aware optimization.

\appendix

\section{Reproducibility Details}\label{reproducibility-details}

Table 2 reports every frozen route used by the final policies. Block
indices are zero-based and identify vision-transformer blocks replaced
by identity. A dash means that budget was not evaluated for that model.

\begin{table}[H]
\centering
\small
\setlength{\tabcolsep}{4pt}
\begin{tabular}{llp{0.19\linewidth}p{0.23\linewidth}p{0.28\linewidth}}
\toprule
Model & Route family & K4 & K6 & K8 \\
\midrule
Qwen & Shared & 3, 9, 20, 28 & 5, 9, 11, 25, 28, 29 & 4, 6, 8, 9, 20, 24, 25, 28 \\
Qwen & Attribute & 4, 8, 9, 22 & 3, 8, 9, 18, 20, 25 & 3, 6, 8, 13, 20, 25, 26, 28 \\
Qwen & Counting & 3, 19, 24, 25 & 4, 5, 10, 20, 21, 29 & 3, 6, 11, 12, 17, 24, 27, 28 \\
Qwen & Object & 3, 5, 10, 19 & 2, 5, 9, 10, 23, 30 & 2, 3, 9, 10, 20, 21, 23, 30 \\
Qwen & OCR & 5, 9, 25, 28 & 4, 5, 9, 26, 27, 28 & 4, 5, 9, 11, 25, 26, 27, 28 \\
Qwen & Spatial & 3, 19, 24, 25 & 3, 5, 12, 20, 22, 26 & 4, 6, 16, 19, 22, 26, 28, 29 \\
\midrule
SmolVLM2 & Shared & 4, 22, 24, 25 & -- & -- \\
SmolVLM2 & Attribute & 5, 11, 22, 25 & -- & -- \\
SmolVLM2 & Counting & 11, 16, 20, 24 & -- & -- \\
SmolVLM2 & Object & 4, 16, 21, 22 & -- & -- \\
SmolVLM2 & OCR & 5, 11, 19, 25 & -- & -- \\
SmolVLM2 & Spatial & 4, 19, 21, 24 & -- & -- \\
\bottomrule
\end{tabular}
\caption{Frozen block-skipping routes. Each cell is one exact fixed-cardinality route, not an ordered execution path.}
\label{tab:frozen-routes}
\end{table}

All searches use seeds 20260715, 20260716, and 20260717. Qwen uses
population 16 and three evaluated generations for K4, K6, and K8;
SmolVLM2 uses population 12 and two evaluated generations for K4. The
Qwen frozen-route and search-configuration SHA-256 hashes are
\nolinkurl{1ca8d3583c6eed93e0375dafc96f90984e8212730be5c2ebcb4be84f9fbfb67c}
and
\nolinkurl{41ba104f2d9df24b67d38e137377de1734e6ada9c7c5f1148c4d3e4afea0170f}.
The corresponding SmolVLM2 hashes are
\nolinkurl{78dccefccc6bf8174d07539830f2014823266d042af9940ac6b9d047a3ef13f4}
and
\nolinkurl{ab27d6e3d8df1bf7407f380e8c4fbe29c5b79ced0a2af6fa00970524fa05f424}.
Source, aggregate evidence, and regeneration scripts are available at
\url{https://github.com/TarunTomar122/vision-pathways} under the
\texttt{arxiv-v1-candidate} tag.

\section*{Acknowledgements}

AI-assisted tools were used for code assistance, experiment
orchestration, figure generation, manuscript organization, and language
editing. No AI system is an author; responsibility for the submitted
work remains with the human author.

\bibliography{references.bib}

\end{document}